# Proactive Intervention to Downtrend Employee Attrition using Artificial Intelligence Techniques

Aasheesh Barvey [a*], Jitin Kapila [b*] & Kumarjit Pathak [c]

**Abstract:** To predict the employee attrition beforehand and to enable management to take individualized preventive action. Using Ensemble classification modeling techniques and Linear Regression. Model could predict over 91% accurate employee prediction, lead-time in separation and individual reasons causing attrition. Prior intimation of employee attrition enables manager to take preventive actions to retain employee or to manage the business consequences of attrition. Once deployed this will model can help in downtrend Employee Attrition, will help manager to manage team more effectively. Model does not cover the natural calamities, and unforeseen events occurring at an individual level like accident, death etc.

## I. -INTRODUCTION

In modern world technology is driving the business and people are becoming more and more important in improving technologies. Thus, for the leading companies keeping their employees motivated and, engaged has become success mantra. There are multiple effects of employee attrition on organization (both tangible and non-tangible) and it's not a surprise that keeping employee attrition under check has become an issue for many organizations. Many managers do have a goal to managing attrition below a set percentage in their appraisals.

There are many adverse impact of employee attrition on organization. Gawali (2009) mentioned soaring employee turnover can lead to the lack of motivation and low morale. Curtis & Wright (2001) indicated that employee attrition also affects customer services and quality. Porter (2011) highlighted financial impacts of employee attrition. Wiley (2011) pointed out that replacement cost of a manager can be twice of the earnings. On the other hand, Lau & Albright (2011) pointed out positive impacts of employee turnover in few cases.

A survey report by Ann Bares (Altura Consulting Group, 2017) pointed out that total employee turnover in 2016 across industries was close to 18%. And it is constantly increasing from 2013 onwards. This clearly shows the magnitude of this business problem

## II -SIMILAR WORK AND LITERATURE SURVEY

As people analytics techniques are getting popular over the period, many organizations had started utilizing employee data to predict likely attrition. Many organizations are still encouraging managers to track life events of their employees to access possible attrition. IBM Watson has done very good work in this field and their sample dataset was used by many researchers including us.

John Yanez, Chief Operating Officer at in his white paper "The Power of Retention: Maximizing Value in Centers of Excellence" has explained benefits of employee downtrend management (lower employee turnover and higher retention rates) for an organization. Cost of attrition, cost of replacement and other parts were calculated and explained very beautifully but that does not cover the attrition prediction part.

Tzur Keren from Otipo has developed a dynamic dashboard to provide an overview of employee to showcase Otipo™ Pulse – Real-Time Workforce Analytics for Organizational Health. This is very useful to understand the challenges in shift based organizations. But that also does not indicate for employees who are on verge of leaving

[a*] Mr. Aasheesh Barvey, Data Scientist professional , Harman , Whitefield, Bangalore ,mail:ashbarvey@gmail.com.
[b*] Mr. Jitin Kapila, Data Scientist professional , Zeta Global , Indiranagar, Bangalore ,mail:Jitin.kapila@outlook.com
[c] Mr. Kumarjit Pathak,  Data Scientist professional , Harman , Whitefield, Bangalore ,mail:Kumarjit.pathak@outlook.com.

[*] major contributor

the organization.

CGI Advanced Analytics team has worked on sample data of ~1500 employees and was able to predict 90% attrition by using "NO FREE LUNCH THEOREM" which is combination of deep learning, neural networks or logistic regression. They had also indicated suggested to map social communications by employees to improve the performance of model. Another limitation is this prediction does not provide what is the "time in hand" to retain the employee.

Matt Dancho et.al [4] recently used machine learning techniques h2o.automl () from "h2o" package to predict employee attrition and used "lime" to understand feature importance to know cause of attrition. The recall of their model output was 62%.

Bhanuprakash et.al [4] has used IBM Watson Dataset and used XGBTree and got accuracy of 89% to predict employee attrition. But he had faced a challenge of high number of false positives (Specificity : 0.5085) . Abbas Heiat, had also used the same dataset used Data mining techniques and got the similar results.

In a nutshell we can summarize that current solution available are limited only to predict the possible % of employee attrition and can also point which employee is going to leave the organization. Still that leaves a gap to know why an employee will leave the organization and how long he will take to really move out.

## III - PROPOSED SOLUTION

As mentioned above current practices and solutions are focusing only on employee attrition part, and good as indicative measures. We felt that business needs actionable item list not just indicative measures. Hence, we took an approach to build a solution which not only predict employee attrition but will also provide action items for his manager or organization to retain the potential attrition case. We had also though to calculate the lead time available with organization to deploy these action items.

Hence the approach we took to build the model to answer the following questions

- Which employees are likely to leave the organization?
- Why are they thinking to leave the organization?
- How much time is left in their final parting off?

As mentioned earlier we had used the same IBM Watson dataset as base dataset. During exploratory analysis and first model iteration using ML classification methods it was observed that model output has limited scope based on variables provided in the dataset. Hence there was a need of feature engineering to make this dataset more meaningful from business perspective.

We have derived many more customized KPIs such as market demand of employee's skills, previous employment history. Market demand of skills of employee was rated on a five-point scale from high to low having five as highest value and 1 as lowest value. We have used industry reports to decile the ratings.

Similarly, previous employment history of an employee can provide number of companies worked, minimum tenure, maximum tenure served in previous organizations. These data points were added using logical variable engineering. This has helped us to predict the tenure of employee in current organization.

Finally, we had factored total 58 variables of dataset into six broad dimensions
- a) Environment Variables within the company (facilities, diversification, profitability of company, etc.)
- b) Financial Variables (Salary, CTC, etc.)
- c) External Variables (Distance between office and home, travel mode etc.
- d) Work related Variables (Business Travel, date of last promotion, change in reporting manager, average working hours etc.)
- e) Legal Variables (Minimum Age for working, Number of scheduled working hours, permission to work / work permit etc.)
- f) Individual Variables (performance ratings, Marital status, No of children, etc.)

Behavior of variables

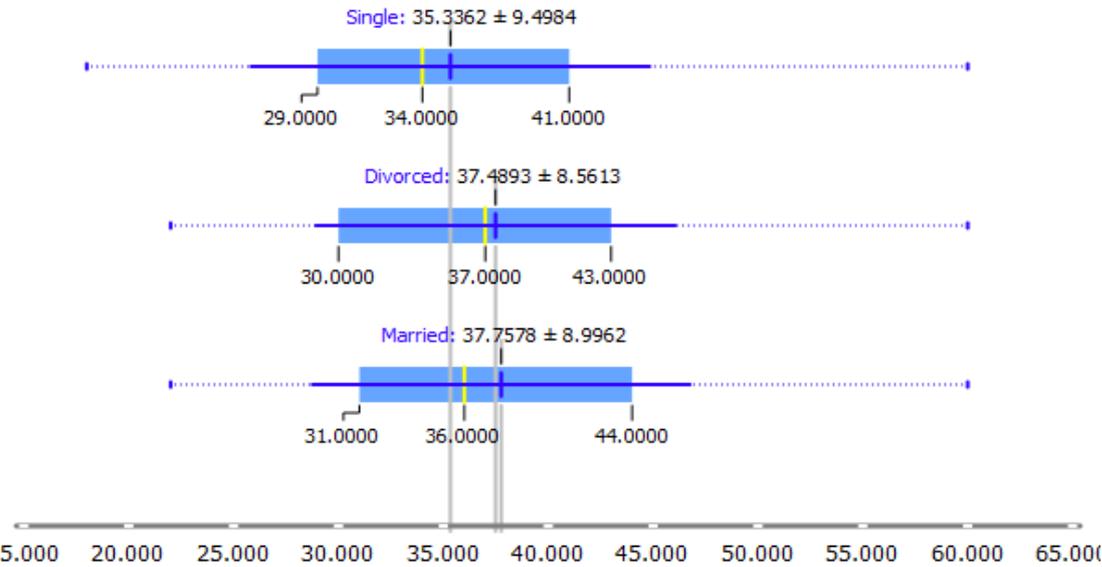

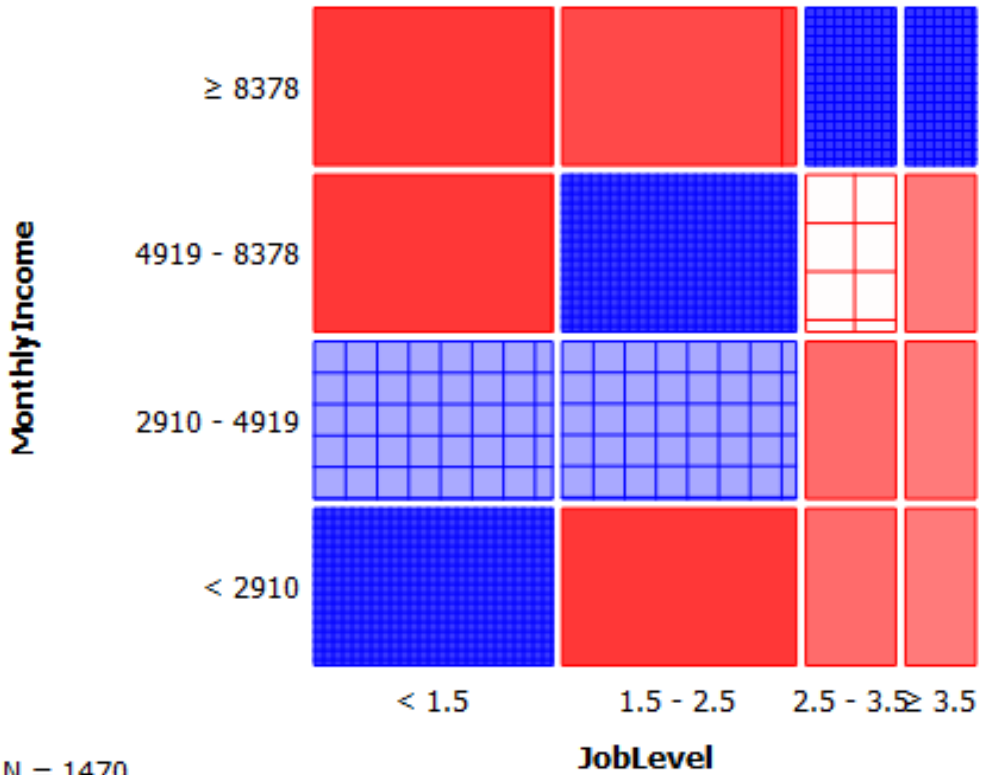

**Modeling techniques and methodologies**

**Artificial Intelligence**

Artificial intelligence (AI) is the intelligence exhibited by machines or software. It is a branch of computer science that deals with helping machines find solutions to complex problems in a more human-like fashion. This generally involves borrowing characteristics from human intelligence and applying them to the design of computer programs. Although primarily thought of as a science, AI is also considered the art of creating machines that perform functions that require intelligence when performed by people.

**Machine Learning**

Machine learning (ML) is the ability of a computer to improve its own performance through the use of software that employs artificial intelligence techniques to mimic the ways by which humans seem to learn. It focuses on the development of computer programs that can teach themselves to change and behave in some improved way when exposed to new data. Machine learning attempts to "learn" by looking for and detecting patterns in the data and adjusting the program's actions accordingly, similar to how humans try to improve understanding through observation, study and repetition.

**Modeling techniques**

For this model we had used ensemble classification methodology to classify between employees who had left the organization and those who are still working in the organization. Post that we had matched patterns of existing employees with those who had already left the organization.
Based on pattern employee attrition was predicted. We have only one data set with us, as per industry practice we split the dataset into two parts 80% was used to train the model and 20% was used as validation dataset.

We got the accuracy 91% in predicting employees who can leave the organization in validation dataset. Since this is a self-learning algorithm accuracy is likely to increase over period of time with other datasets.

We had used linear regression to predict the total tenure of each employees in the organization, this way we were able to get the "lead time" available with management for the employees who can leave the organization in the future.

$$LT = TTL - CT$$

(LT = lead time, TLT= total tenure length and CT current tenure with organization).

The highlighting part is that model's prediction range which takes care of capturing the likely attrition cases in near future (within couple of weeks) and well as long run attrition cases (more than a year)

We also did key driver analysis at each employee to assess the likely reasons for all employees who may leave the organization in future. This was performed on output of ensemble algorithm.

## IV - IMPLEMENTATION STRATEGY

Currently Model is built in R and output is portrayed using Shiny framework. Once deployed in any organization model can fetch data from HR platforms (SAP, Oracle, etc) and run on regular intervals to predict likely attrition cases, Lead time for each attrition case and variables responsible for the attrition.

The output is available in a user-friendly dashboard. This can be further customized as per the requirement of organization. Before deployment, one of the key requirement area will remain user level access and calibration of model.

Since every organization is different from another one, we believe that model calibration will be needed to address that difference.

## V - POTENTIAL BENEFITS

Organizations keep on running various initiatives to increase employee engagement. The proposed solution can work as lead indicator to highlight areas where employee engagement needed to be addressed. This will help to down trend the employee attrition with proactive actions.

## VI FUTURE SCOPE

The same model can also be used to predict the fitment of new employee in the organization.
If there are multiple deserving candidates available for one position this model can predict the most suitable candidate in terms of cultural fitment and longest tenure in new organization.

Any change in the working condition of employee (like manager change etc) can be assessed in advance and based on merits, management can take the decision.

## VI LIMITATIONS OF STUDY

The model does not cover any natural calamity, war, political unrest or any unforeseen events may occur in any employees life like accident, etc. The scope of model is restricted within the variables feuded in the algorithm and purity of available data.

**Model consumption and business planning dashboard:**

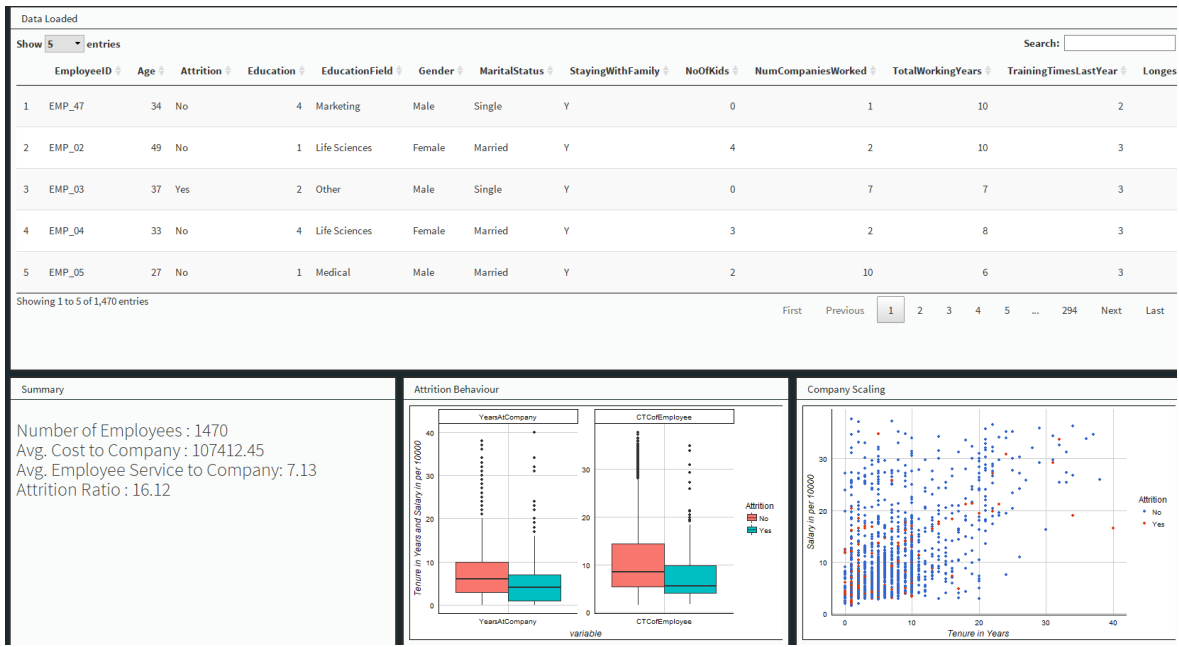

This part of dashboard provide overall overview of Employee Engagement level at organizational level at the same time also provide details of each employee. It indicates Average CTC per employee, total employee counts, attrition ratio at organizational level.

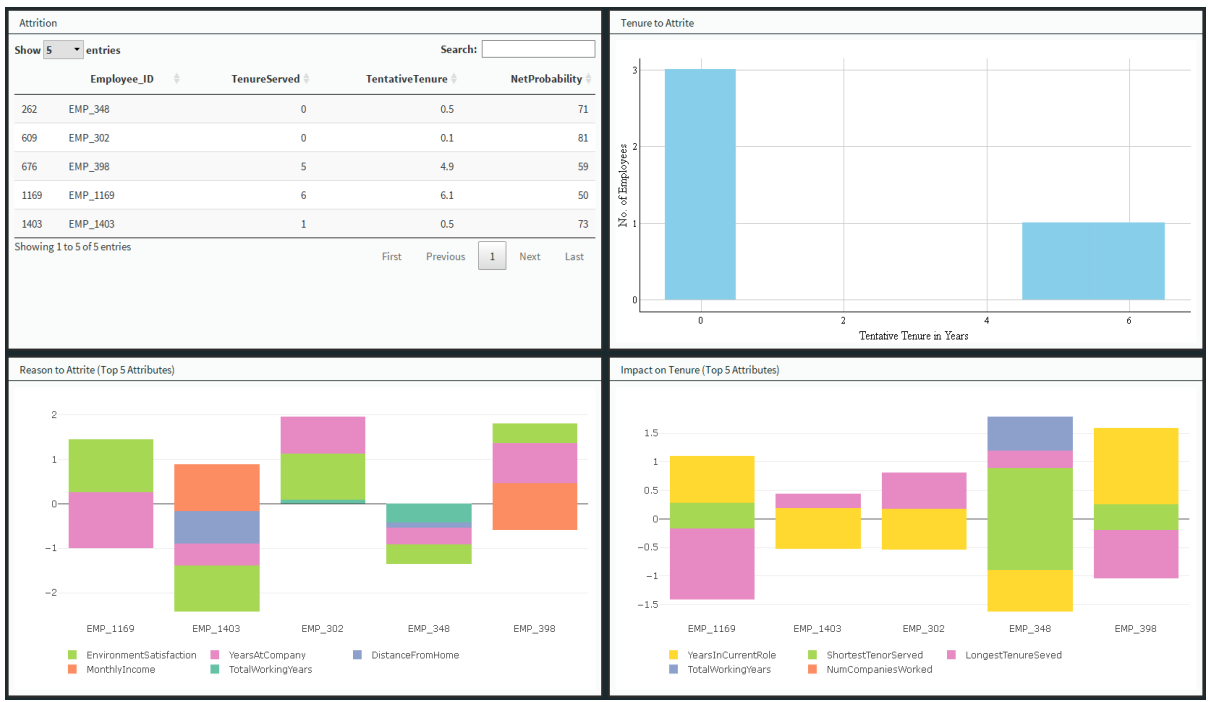

This dashboard focuses on attrition prone employees and it provides deep dive of attrition that is top reasons propelling to attrition, and which variables are affecting the tenure of service. It also provides the lead-time organization have to facilitate preventive actions.

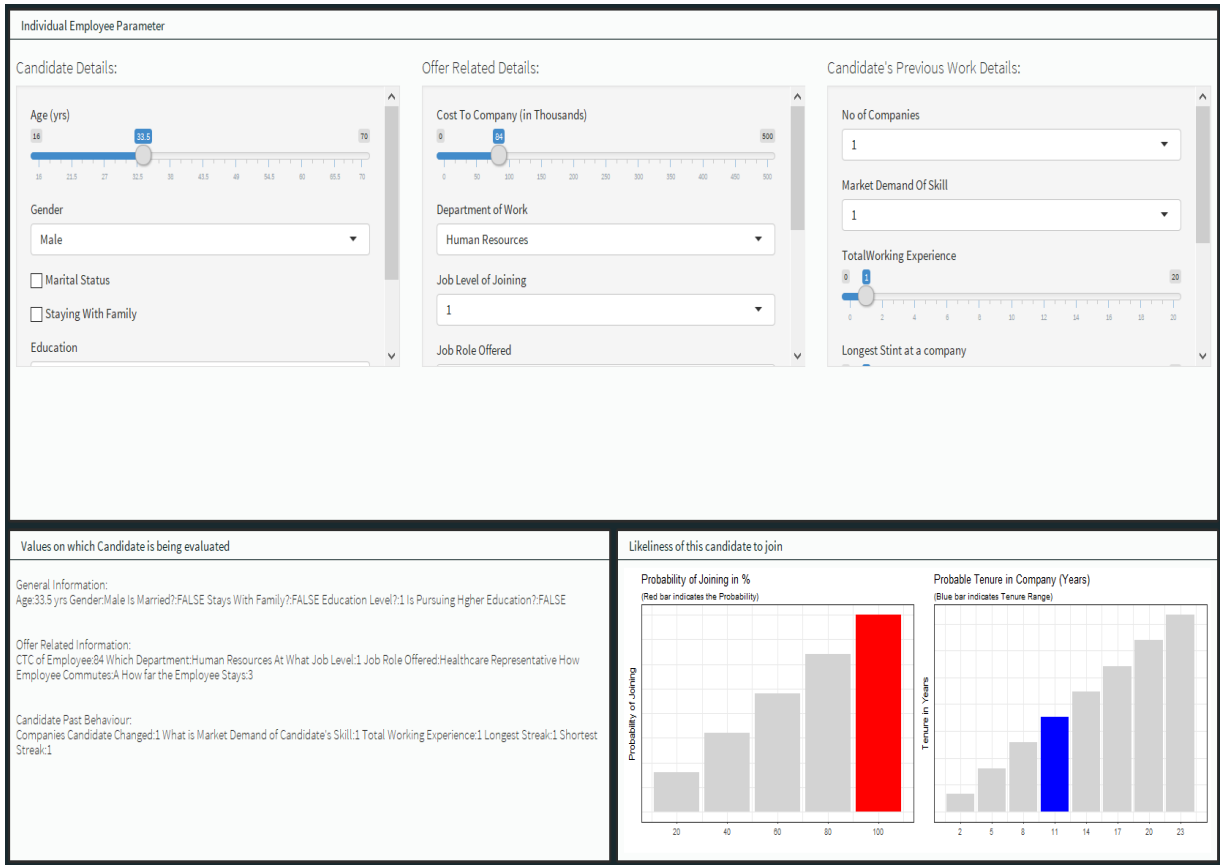

This dashboard provides an screening option for Organization to check for cultural fitment of probable employee and his likelihood of joining the organization. This may come handy for taking a final evaluation call between 2 or more deserving candidates for one post.